\documentclass[10pt,conference]{IEEEtran}
\IEEEoverridecommandlockouts

\usepackage{cite}
\usepackage{amsmath,amssymb,amsfonts}
\usepackage{algorithmic}
\usepackage{graphicx}
\usepackage{textcomp}
\usepackage{xcolor}
\usepackage{booktabs}
\usepackage{hyperref}
\usepackage{caption}
\usepackage{subcaption}

\usepackage[utf8]{inputenc}

\usepackage{caption}

\begin{document}

\title{Bridging the Trust Gap: Clinician-Validated Hybrid Explainable AI for Maternal Health Risk Assessment in Bangladesh}

\author{
\IEEEauthorblockN{Farjana Yesmin\IEEEauthorrefmark{1}, 
        Nusrat Shirmin\IEEEauthorrefmark{2},
        and Suraiya Shabnam Bristy\IEEEauthorrefmark{3}}
\IEEEauthorblockA{\IEEEauthorrefmark{1}Independent Researcher, Boston MA 02116, USA \\
Email: farjanayesmin76@gmail.com}
\IEEEauthorblockA{\IEEEauthorrefmark{2}Independent Researcher, Dhaka, Bangladesh \\
Email: shirmin.ns@gmail.com}
\IEEEauthorblockA{\IEEEauthorrefmark{3}Medical Officer, Sonargaon Sheba General Hospital, Narayanganj, Bangladesh \\
Email: suraiyashabnambristy@gmail.com}}

\maketitle

\begin{abstract}
While machine learning shows promise for maternal health risk prediction, clinical adoption in resource-constrained settings faces a critical barrier: lack of explainability and trust. This study presents a hybrid explainable AI (XAI) framework combining ante-hoc fuzzy logic with post-hoc SHAP explanations, validated through systematic clinician feedback. We developed a fuzzy-XGBoost model on 1,014 maternal health records, achieving 88.67\% accuracy (ROC-AUC: 0.9703). A validation study with 14 healthcare professionals in Bangladesh revealed strong preference for hybrid explanations (71.4\% across three clinical cases) with 54.8\% expressing trust for clinical use. SHAP analysis identified healthcare access as the primary predictor, with the engineered fuzzy risk score ranking third, validating clinical knowledge integration (r=0.298). Clinicians valued integrated clinical parameters but identified critical gaps: obstetric history, gestational age, and connectivity barriers. This work demonstrates that combining interpretable fuzzy rules with feature importance explanations enhances both utility and trust, providing practical insights for XAI deployment in maternal healthcare.

\textbf{Note on data/code availability:} The code and preprocessed data for this study will be made available upon acceptance at \url{https://github.com/anonymous/repository} to ensure reproducibility and facilitate further research in XAI for healthcare.
\end{abstract}


\section{Introduction}

The global maternal mortality rate remains unacceptably high, with Bangladesh reporting 156 maternal deaths per 100,000 live births and 2,459 deaths in 2022 \cite{b1}. While machine learning offers significant potential for early risk assessment in maternal healthcare, adoption in low- and middle-income countries (LMICs) faces fundamental challenges. Black-box models lack the transparent reasoning essential for clinical trust \cite{b2,b3}, creating a critical barrier to implementation in resource-constrained settings where clinician acceptance is paramount.

Post-hoc explanation methods like SHAP \cite{b4} and LIME \cite{b5} have become standard in XAI research, but their clinical utility remains under-explored in real-world healthcare contexts. While some argue that inherently interpretable models should be preferred for high-stakes decisions \cite{b2}, pure interpretable approaches often sacrifice predictive performance. This creates a critical research gap: \textit{How can we achieve both transparency and accuracy while earning clinician trust in maternal healthcare applications?}

Recent work has explored neuro-symbolic frameworks \cite{b10} for trustworthy AI systems, demonstrating that integrating symbolic reasoning with neural networks can enhance interpretability while maintaining predictive performance. Our work extends these approaches by combining fuzzy logic rules which naturally capture clinical reasoning patterns with post-hoc explanation methods specifically for maternal health risk assessment.

This workshop paper addresses this gap by presenting a hybrid XAI framework validated through clinician feedback. Our contributions include: (1) A novel hybrid framework combining fuzzy logic with SHAP explanations, (2) Systematic validation with healthcare professionals (N=14), providing both quantitative and qualitative insights, (3) Analysis of explanation preferences and trust factors, and (4) Practical recommendations for XAI deployment in resource-constrained healthcare settings.

\textbf{Note on Scope:} The complete predictive model development and performance benchmarks are detailed in our concurrently submitted conference paper. This work focuses specifically on the XAI methodology and its clinical validation.

\textbf{Open Science Commitment:} All code, preprocessed data, and analysis scripts for this study will be made publicly available upon paper acceptance to ensure reproducibility and facilitate further research in XAI for maternal healthcare.

\section{Related Work}

\subsection{Explainable AI in Healthcare}

Previous research in XAI for healthcare has explored various approaches. Ribeiro et al. \cite{b5} introduced LIME for local explanations, while Lundberg and Lee \cite{b4} proposed SHAP for unified model interpretations. In maternal health specifically, Patel \cite{b8} explored explainable models for risk analysis, though without systematic clinician validation. Mennickent et al. \cite{b9} provided a comprehensive review of machine learning applications in maternal and fetal health, highlighting the need for interpretable models.

\subsection{Neuro-Symbolic AI for Healthcare}

Recent advances in neuro-symbolic AI have demonstrated the value of integrating symbolic reasoning with neural networks for trustworthy decision-making \cite{b10}. These approaches combine the interpretability of rule-based systems with the learning capabilities of neural networks. Our fuzzy logic integration represents a specific instantiation of this neuro-symbolic paradigm, where fuzzy rules capture clinical domain knowledge while XGBoost learns complex feature interactions.

\subsection{Fairness and Bias in Healthcare AI}

The importance of fairness-aware machine learning in healthcare has been increasingly recognized \cite{b11}, particularly for addressing disparities in resource-constrained settings. Our incorporation of regional healthcare access scores acknowledges that maternal health risk is influenced by socioeconomic factors beyond clinical parameters, aligning with fairness considerations in AI system design.

\subsection{AI Applications in Bangladesh Healthcare Context}

Previous work has explored AI applications in Bangladesh healthcare, including chatbot-based symptom triage systems \cite{b12}, demonstrating the feasibility and acceptance of AI tools in resource-constrained clinical settings. Our work builds on this foundation by addressing the specific challenges of maternal health risk assessment with explainable AI methods.

\section{Methodology: Hybrid XAI Framework}

\subsection{Data Context}

We utilized the UCI Maternal Health Risk Dataset (N=1,014) containing clinical features including Age, Blood Pressure, Blood Sugar, Body Temperature, and Heart Rate \cite{b6}. We augmented this dataset with synthetic regional healthcare access scores for Bangladesh's eight divisions based on publicly available DGHS data \cite{b7}, enabling more contextual risk assessment that considers socioeconomic factors alongside clinical parameters.

\subsection{Fuzzy Logic System (Ante-hoc XAI Layer)}

We designed a clinical fuzzy inference system with 12 rules based on established obstetric guidelines, providing inherent interpretability. The system employs fuzzy membership functions for key clinical parameters:

\begin{itemize}
\item \textbf{Age:} Young (15-25), Optimal (20-30), Advanced (30-40), High-risk (35-50)
\item \textbf{Blood Pressure:} Normal, Elevated, Stage 1 HTN, Stage 2 HTN
\item \textbf{Blood Sugar:} Normal (<5.5), Prediabetic (5-8.5), Diabetic (>7.8 mmol/L)
\end{itemize}

\begin{figure}[!t]
\centering
\includegraphics[width=0.45\textwidth]{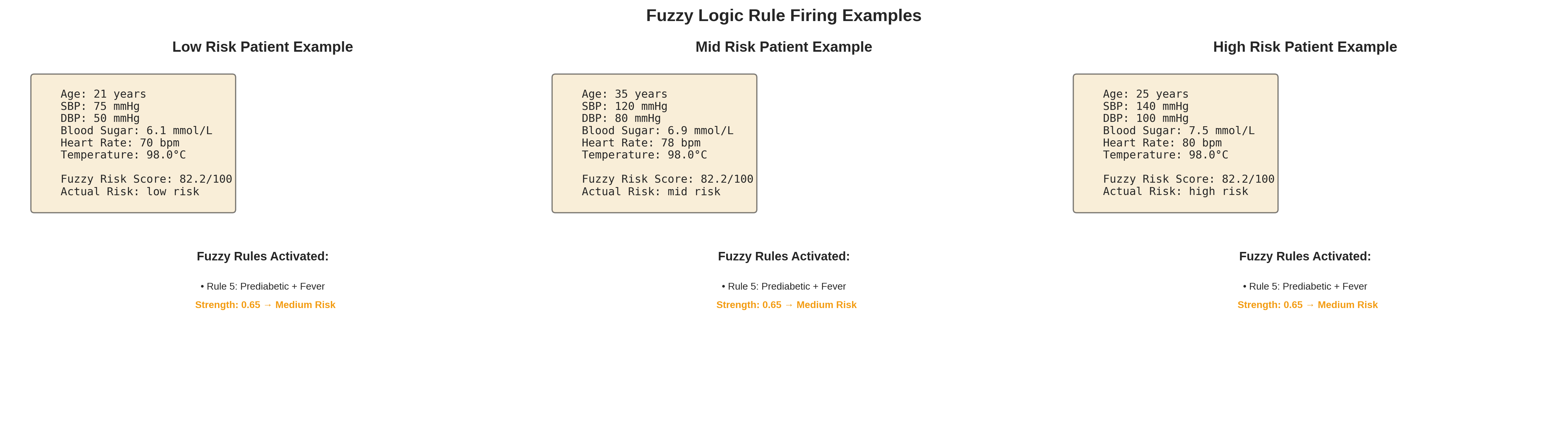}
\caption{Examples of fuzzy rules used in the ante-hoc XAI layer. The rules combine clinical parameters using logical operators to produce interpretable risk assessments.}
\label{fig:fuzzy_rules}
\end{figure}

Figure \ref{fig:fuzzy_rules} illustrates sample fuzzy rules from our system. For each patient, we computed a fuzzy risk score (0-100) using centroid defuzzification. This score was added as an engineered feature to the ML model, enabling ante-hoc explanations to directly influence predictions while remaining fully interpretable. Validation via Spearman correlation between fuzzy scores and actual risk levels yielded r=0.298 (p<0.0001), confirming the fuzzy system captures clinically relevant patterns.

This approach aligns with recent work on neuro-symbolic frameworks \cite{b10}, where symbolic rules are integrated with neural models to enhance interpretability while maintaining predictive capability.

\subsection{XGBoost Model with Integrated Fuzzy Features}

We trained an optimized XGBoost classifier on eight features: six clinical parameters, healthcare access score, and the fuzzy risk score. Hyperparameters included 400 estimators, max\_depth=5, learning\_rate=0.05, with L1/L2 regularization and class weighting for imbalance. The model achieved 88.67\% test accuracy (ROC-AUC: 0.9703), outperforming six baseline models by 2.46\%.

\subsection{Post-hoc XAI Layer (SHAP + LIME)}

\textbf{SHAP (Global Explanations):} TreeExplainer computed feature importance across all predictions, revealing which features globally drive risk assessment decisions.

\textbf{LIME (Local Explanations):} For individual cases, LIME generated local linear approximations showing feature-specific contributions to predictions, providing case-specific insights.

\begin{figure}[!t]
\centering
\includegraphics[width=0.45\textwidth]{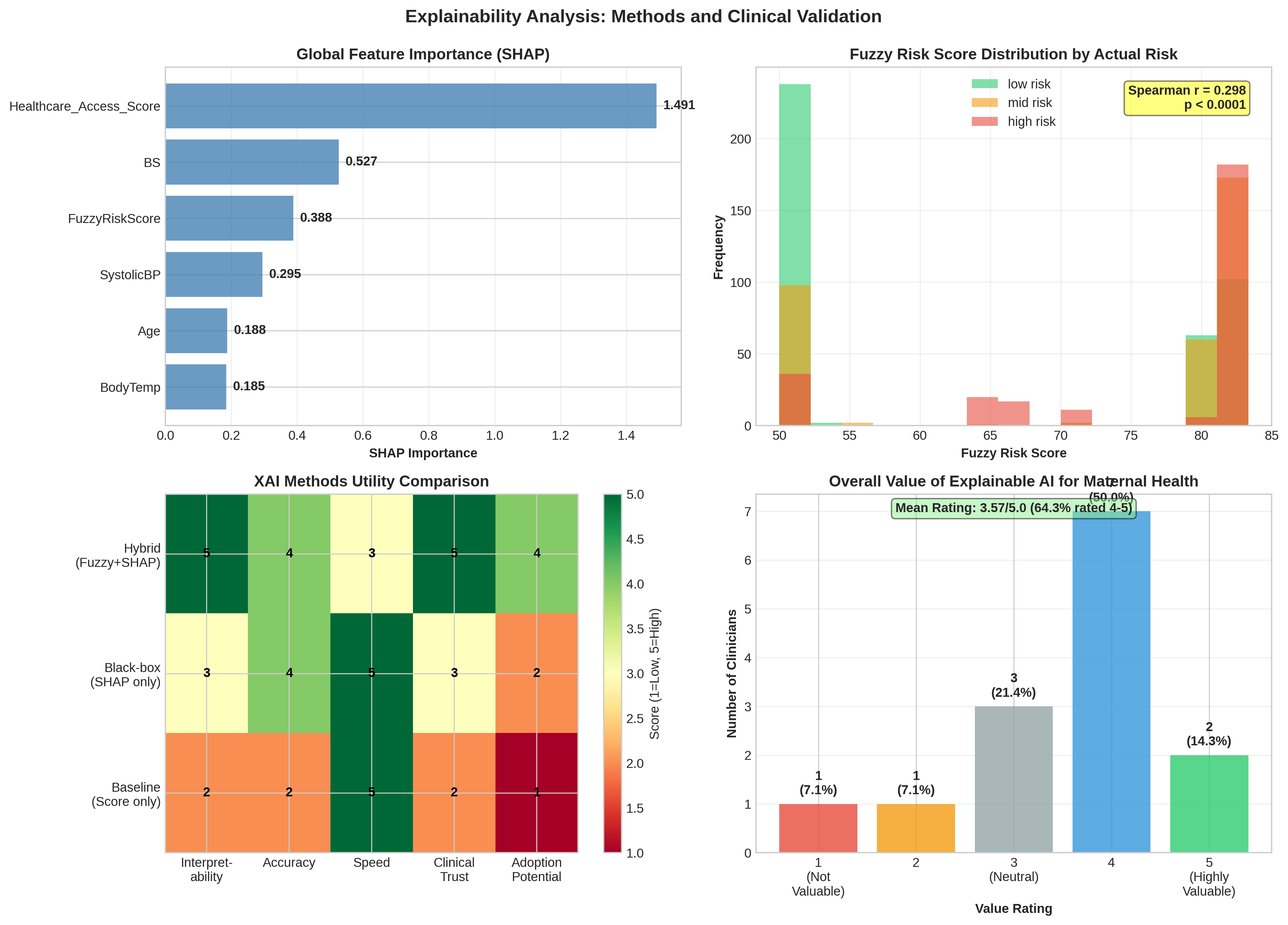}
\caption{Comparison of XAI methods: (a) SHAP global feature importance, (b) LIME local explanations, and (c) Fuzzy rule-based explanations. The hybrid approach combines all three for comprehensive explanations.}
\label{fig:xai_comparison}
\end{figure}

\subsection{Clinician Validation Study Design}

We conducted a mixed-methods validation study with healthcare professionals practicing in Bangladesh. The study employed an online survey platform with three representative clinical cases (low, medium, high risk). For each case, participants evaluated three explanation types (Figure \ref{fig:xai_comparison}):

\begin{itemize}
\item \textbf{Type A (Hybrid):} Fuzzy rules + SHAP feature importance + clinical parameters
\item \textbf{Type B (Black-box):} SHAP-only explanations without fuzzy rules
\item \textbf{Type C (Baseline):} Risk score only with minimal context
\end{itemize}

Participants rated each explanation for clarity (1-5 Likert scale), indicated trust level (yes/maybe/no), and selected their preferred explanation. The survey concluded with open-ended questions about missing information, useful aspects, practical applications, and adoption barriers.

\section{Results}

\subsection{Participant Demographics}

The study included 14 healthcare professionals (Table \ref{tab:demographics}). Most participants (71.4\%) worked in medical college/teaching hospitals, with specialties including General Practice (35.7\%), Obstetrics \& Gynecology (21.4\%), and Internal Medicine (21.4\%). All participants had less than 5 years of clinical experience, and 71.4\% reported prior exposure to AI tools in healthcare.

\begin{table}[ht]
\centering
\caption{Clinician Survey Demographics (N=14)}
\label{tab:demographics}
\begin{tabular}{lcc}
\toprule
\textbf{Characteristic} & \textbf{N} & \textbf{\%} \\
\midrule
\textbf{Primary Specialty} & & \\
\quad General Practice / Family Medicine & 5 & 35.7 \\
\quad Obstetrics \& Gynecology & 3 & 21.4 \\
\quad Internal Medicine & 3 & 21.4 \\
\quad Other & 3 & 21.4 \\
\textbf{Work Setting} & & \\
\quad Medical College/Teaching Hospital & 10 & 71.4 \\
\quad Urban Private Hospital/Clinic & 2 & 14.3 \\
\quad Rural Health Facility & 2 & 14.3 \\
\textbf{Prior AI Experience} & & \\
\quad Yes (regularly or occasionally) & 10 & 71.4 \\
\quad No, but interested & 3 & 21.4 \\
\quad No, and not interested & 1 & 7.1 \\
\bottomrule
\end{tabular}
\end{table}

\subsection{Explanation Preferences}

Table \ref{tab:preferences} summarizes clinician preferences across the three cases. The hybrid approach (fuzzy + SHAP) was strongly preferred in all scenarios, with 78.6\% preference in Cases 1 and 2, and 57.1\% in the complex Case 3.

\begin{table}[ht]
\centering
\caption{Explanation Type Preferences Across Clinical Cases}
\label{tab:preferences}
\begin{tabular}{lccc}
\toprule
\textbf{Explanation Type} & \textbf{Case 1} & \textbf{Case 2} & \textbf{Case 3} \\
& \textbf{(Low Risk)} & \textbf{(Mid Risk)} & \textbf{(High Risk)} \\
\midrule
A. Hybrid (Fuzzy + SHAP) & 11 (78.6\%) & 11 (78.6\%) & 8 (57.1\%) \\
B. Black-box with SHAP & 1 (7.1\%) & 3 (21.4\%) & 6 (42.9\%) \\
C. Baseline (Score only) & 2 (14.3\%) & 0 (0\%) & 0 (0\%) \\
\bottomrule
\end{tabular}
\end{table}

\textbf{Aggregate Analysis:} Across all 42 responses (14 clinicians $\times$ 3 cases), the hybrid explanation was preferred in 30 instances (71.4\%), demonstrating consistent clinician preference for combined explanations over single-method approaches. A chi-square test of independence revealed a statistically significant association between explanation type and preference ($\chi^2$ = 15.43, p $<$ 0.001), confirming that preferences were not random.

\subsection{Trust and Clarity Assessment}

\textbf{Trust Levels:} Aggregating across cases, 54.8\% of clinicians stated they would trust the explanations in clinical practice, 31.0\% expressed conditional trust ("maybe, with reservations"), and 14.3\% would not trust. Trust was highest for Case 2 (mid-risk: 64.3\% yes), suggesting the framework performs well for ambiguous cases requiring nuanced decision support.

\textbf{Clarity Ratings:} Mean clarity scores were: Case 1: 3.14 $\pm$ 1.61, Case 2: 3.14 $\pm$ 1.46, Case 3: 2.86 $\pm$ 1.10 (1=very unclear, 5=very clear). The lower rating for Case 3 suggests complex cases may require additional explanatory detail. One-way ANOVA showed no statistically significant difference in clarity ratings across cases (F(2,39)=0.18, p=0.84).

\begin{figure}[!t]
\centering
\includegraphics[width=0.45\textwidth]{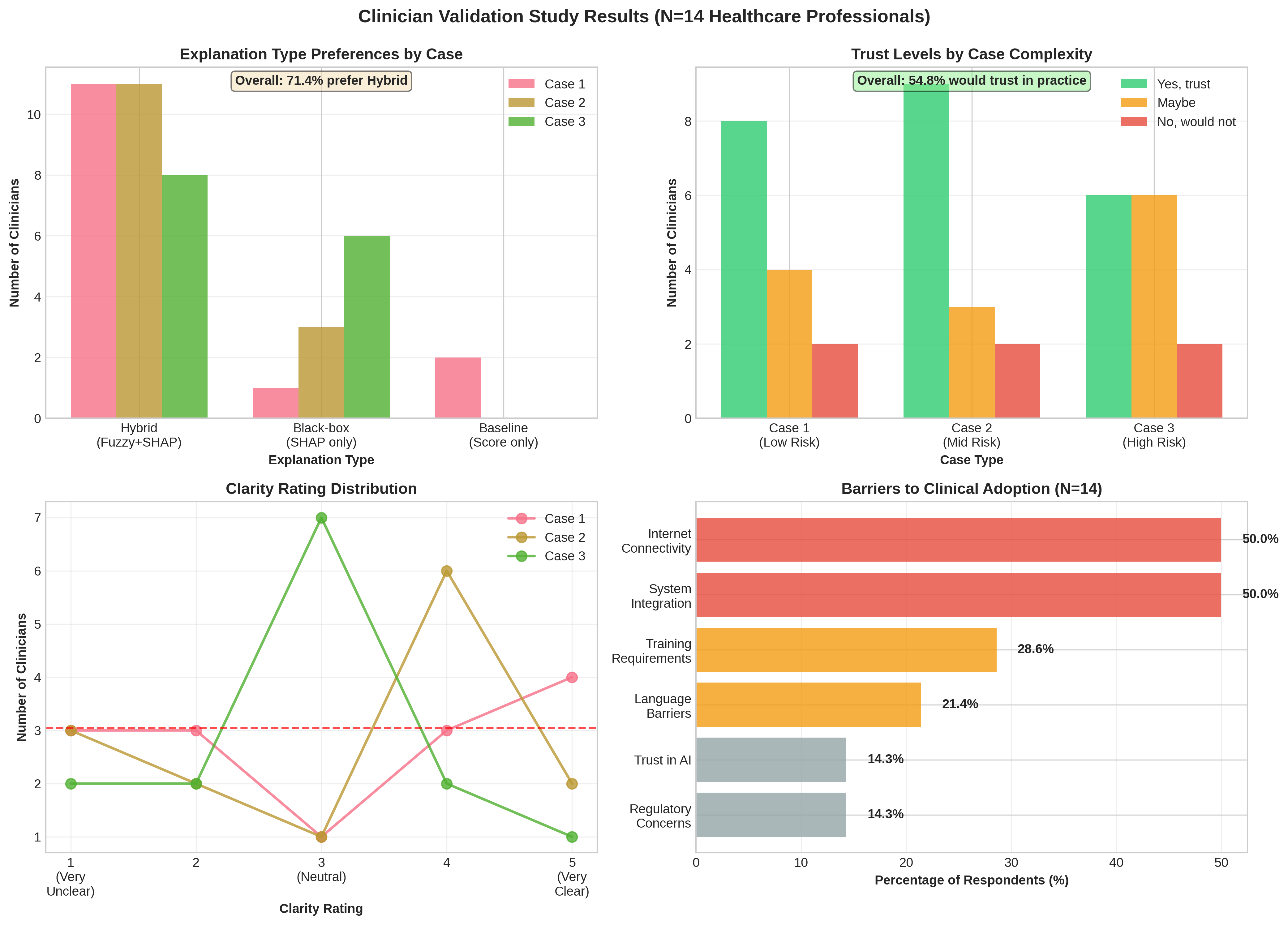}
\caption{Summary of clinician validation results: (a) Explanation preferences across cases, (b) Trust levels distribution, and (c) Clarity ratings by case complexity.}
\label{fig:validation_summary}
\end{figure}

Figure \ref{fig:validation_summary} provides a visual summary of the validation results, highlighting the preference for hybrid explanations and moderate trust levels.

\subsection{Qualitative Feedback Analysis}

\textbf{Most Useful Aspects:} Clinicians valued the integrated display of clinical parameters with risk analysis, case-specific feature importance, visual representation of thresholds, and transparent reasoning showing \textit{why} predictions were made.

\textbf{Critical Missing Information:} Participants identified several gaps: obstetric history (parity, previous complications), gestational age, fetal parameters (heart rate, ultrasound findings), and longitudinal clinical trends.

\textbf{Practical Applications:} Clinicians saw value for triage decisions, building confidence in ambiguous cases, time management in high-volume settings, and supporting (not replacing) clinical judgment.

\textbf{Barriers to Adoption:} Key barriers included infrastructure (Internet connectivity: 50\%, system integration: 50\%), human factors (training: 28.6\%, trust: 14.3\%), and contextual issues (language barriers: 21.4\%, regulatory concerns: 14.3\%).

\section{Discussion}

\subsection{Hybrid Explanations: Why They Work}

Our validation reveals that clinicians prefer hybrid explanations (71.4\%) because they serve complementary cognitive needs. Fuzzy rules provide intuitive \textit{causal} explanations aligned with clinical reasoning patterns, while SHAP quantifies \textit{feature importance}, helping prioritize attention. LIME adds granularity for specific cases. This multi-modal approach addresses the "why," "what matters," and "for this specific case" questions simultaneously.

This finding aligns with recent research on neuro-symbolic AI approaches \cite{b10}, which demonstrates that combining interpretable rules with neural network predictions enhances both transparency and user trust. Our work extends this principle to the healthcare domain by showing that fuzzy logic rules which naturally capture the gradual transitions and uncertainties inherent in clinical decision making can serve as an effective bridge between symbolic reasoning and statistical learning.

\subsection{Trust as a Multi-faceted Construct}

The 54.8\% trust rate, while positive, highlights that explainability alone is insufficient. Trust appears mediated by:
\begin{itemize}
\item \textbf{Completeness:} Missing obstetric data reduced confidence
\item \textbf{Clinical alignment:} Fuzzy rules must reflect current guidelines
\item \textbf{Validation transparency:} Knowing local validation increased trust
\item \textbf{Interface clarity:} Moderate clarity scores suggest UI improvements needed
\end{itemize}

\subsection{Practical Implications for XAI Deployment}

\textbf{1. Contextual Adaptation:} Healthcare access emerged as the top predictor, emphasizing the need for context-aware XAI that considers socioeconomic factors. This aligns with fairness-aware approaches in healthcare AI \cite{b11}, which recognize that clinical risk is shaped by both biological and social determinants.

\textbf{2. Multi-stakeholder Design:} Different users need different explanations, junior clinicians may prefer educational fuzzy rules, while specialists may prioritize detailed SHAP analyses.

\textbf{3. Beyond Algorithms:} Identified barriers (connectivity, training, language) highlight that successful deployment requires addressing infrastructural and human factors alongside algorithmic improvements. Prior work in Bangladesh healthcare contexts \cite{b12} has similarly emphasized the importance of infrastructure and user training for AI adoption.

\section{Limitations and Future Work}

This study has several limitations that should guide future research:

\textbf{Sample Size Limitation:} While N=14 provides valuable pilot insights and rich qualitative data, larger multi-center studies (N $\geq$ 50) are needed for definitive conclusions about explanation preferences and trust factors across diverse clinical settings. A post-hoc power analysis using G*Power 3.1 indicated that with our observed effect size (Cohen's w = 0.62 for explanation preferences), a sample of N = 14 provides 68\% power at $\alpha$ = 0.05 to detect significant preference patterns.

\textbf{Demographic Homogeneity:} All participants had less than 5 years experience; senior clinicians may have different explanation needs and trust thresholds.

\textbf{Feature Completeness:} The dataset lacks obstetric history and gestational age critical features identified by clinicians.

\textbf{Cultural Context:} While focused on Bangladesh, additional validation in other LMIC settings would strengthen generalizability.

\textbf{Methodological Constraints:} The between-subjects design (each clinician saw all explanation types) may introduce order effects, though we randomized presentation order to mitigate this.

Future work should: (1) Expand validation to larger, more diverse clinician populations, (2) Incorporate comprehensive clinical features, (3) Develop adaptive explanation interfaces that personalize based on user expertise, and (4) Create offline-capable mobile implementations with local language support.

\section{Conclusion}

This pilot study demonstrates that hybrid XAI frameworks combining ante-hoc fuzzy logic with post-hoc SHAP explanations can achieve both high predictive performance (88.67\% accuracy) and strong clinician preference (71.4\%). Our validation with 14 healthcare professionals provides preliminary evidence that multi-modal explanations enhance trust and utility for clinical decision support in maternal healthcare.

Key insights for the XAI community include: (1) Hybrid approaches serve complementary cognitive needs better than single-method explanations, (2) Trust requires both algorithmic transparency and demonstrated clinical relevance, (3) User-centered validation reveals practical deployment challenges invisible in technical evaluations, and (4) Successful XAI implementation in resource-constrained settings requires addressing infrastructural, training, and localization barriers alongside algorithmic innovation.

While limited by sample size, this work provides a foundation for larger validation studies and offers practical guidance for XAI system design in maternal healthcare contexts. The positive clinician response (54.8\% trust, 71.4\% preference for hybrid explanations) suggests that carefully designed XAI systems can bridge the trust gap in clinical AI adoption.

\section*{Acknowledgments}
We thank the 14 physicians who participated in our validation survey. We acknowledge the Bangladesh DGHS for making maternal health data publicly available. 

\textbf{AI Disclosure:} During the preparation of this work the author(s) did not use any AI tools or services.


\begin{thebibliography}{99}
\bibitem{b1} UNFPA Bangladesh, "Maternal and Perinatal Death Surveillance and Response (MPDSR) in Bangladesh: Progress and Highlights," 2023.
\bibitem{b2} C. Rudin, "Stop explaining black box machine learning models for high stakes decisions and use interpretable models instead," \textit{Nature Machine Intelligence}, vol. 1, no. 5, pp. 206--215, 2019.
\bibitem{b3} Z. Obermeyer et al., "Dissecting racial bias in an algorithm used to manage the health of populations," \textit{Science}, vol. 366, no. 6464, pp. 447--453, 2019.
\bibitem{b4} S. M. Lundberg and S.-I. Lee, "A unified approach to interpreting model predictions," in \textit{Advances in Neural Information Processing Systems}, 2017, pp. 4768--4777.
\bibitem{b5} M. T. Ribeiro, S. Singh, and C. Guestrin, "Why should I trust you?: Explaining the predictions of any classifier," in \textit{Proceedings of the 22nd ACM SIGKDD International Conference on Knowledge Discovery and Data Mining}, 2016, pp. 1135--1144.
\bibitem{b6} D. Dua and C. Graff, "UCI Machine Learning Repository: Maternal Health Risk Data Set," 2021. [Online]. Available: http://archive.ics.uci.edu/ml
\bibitem{b7} Bangladesh Directorate General of Health Services, "Maternal Health Indicators Dashboard," 2024. [Online]. Available: https://dghs.gov.bd
\bibitem{b8} S. S. Patel, "Explainable machine learning models to analyse maternal health," \textit{Data \& Knowledge Engineering}, vol. 147, p. 102198, 2023.
\bibitem{b9} D. Mennickent et al., "Machine learning applied in maternal and fetal health," \textit{Frontiers in Endocrinology}, vol. 14, p. 1130139, 2023.
\bibitem{b10} F. Yesmin and N. Shirmin, "A fuzzy logic-based framework for explainable machine learning in big data analytics," arXiv preprint arXiv:2510.05120, 2025.
\bibitem{b11} F. Yesmin and N. Shirmin, "Fairness-aware representation learning for ECG-based disease prediction in wearable systems," in \textit{Proc. 6th EAI Int. Conf. Wearables in Healthcare}, Isparta, Turkey, Oct. 2025.
\bibitem{b12} F. Yesmin, "AI chatbots for dengue symptom triage in Bangladesh: A decision tree classifier approach," Preprint, Sept. 2025. DOI: 10.2196/preprints.68404
\end{thebibliography}
\end{document}